\documentclass[letterpaper, 10 pt, conference]{ieeeconf}  

\IEEEoverridecommandlockouts                              

\usepackage{graphics} 
\usepackage{epsfig} 
\usepackage{amsmath} 
\usepackage{amssymb}  
\usepackage{pbalance}

\title{\LARGE \bf
Toward Improving fNIRS Classification: A Study on Activation Functions in Deep Neural Architectures
}

\author{Behtom Adeli$^{1}$, John P. McLinden$^{1}$, PhD. Pankaj Pandey$^{2}$, PhD. \\ Ming Shao$^{3}$, PhD. \textit{Senior Member, IEEE,} Yalda Shahriari$^{1}$, PhD.$^{*}$  \textit{Senior Member, IEEE} \\ $^{*}$Corresponding author; Email: yalda\_shahriari@uri.edu
\thanks{*This work was supported by the National Science Foundation.}
\thanks{$^{1}$Department of Electrical, Computer, \& Biomedical Engineering, University of Rhode Island, Kingston, RI, USA.
}%
\thanks{$^{2}$St. Jude Children's Research Hospital, Memphis, TN, USA.
       }%
\thanks{$^{3}$Miner School of Computer and Information Sciences at University of Massachusetts Lowell, MA, USA.
       }%
}%

\begin{document}

\maketitle
\thispagestyle{empty}
\pagestyle{empty}

\begin{abstract}
Activation functions are critical to the performance of deep neural networks, particularly in domains such as functional near-infrared spectroscopy (fNIRS), where nonlinearity, low signal-to-noise ratio (SNR), and signal variability poses significant challenges to model accuracy. However, the impact of activation functions on deep learning (DL) performance in the fNIRS domain remains underexplored and lacks systematic investigation in the current literature. This study evaluates a range of conventional and field-specific activation functions for fNIRS classification tasks using multiple deep learning architectures, including the domain-specific fNIRSNet, AbsoluteNet, MDNN, and shallowConvNet (as the baseline), all tested on a single dataset recorded during an auditory task. To ensure fair a comparison, all networks were trained and tested using standardized preprocessing and consistent training parameters. The results show that symmetrical activation functions such as Tanh and the Absolute value function Abs(x) can outperform commonly used functions like the Rectified Linear Unit (ReLU), depending on the architecture. Additionally, a focused analysis of the role of symmetry was conducted using a Modified Absolute Function (MAF), with results further supporting the effectiveness of symmetrical activation functions on performance gains. These findings underscore the importance of selecting proper activation functions that align with the signal characteristics of fNIRS data.
\newline


\end{abstract}

\section{INTRODUCTION}
Convolutional neural networks (CNNs) have achieved state-of-the-art performance in diverse domains, from computer vision to signal processing. One of the most important components of each CNN is the activation function, which is the main operator and greatly influences the network's accuracy, convergence, and generalizability \cite{Liu, Eastmond}. Different studies have explored the effect of activation functions on the performance of CNNs in other fields ~\cite{EmanuelCNN,dubeyactivation,rawat, wangactivation}. For example, Mehta et al. studied different activations for a simple Artificial Neural Network (ANN) on electroencephalography (EEG) signal classification and found that functions such as the Exponential Linear Unit (ELU) and leaky ReLU can achieve greater accuracy than pure Rectified Linear Unit (ReLU) on their small dataset ~\cite{Mehta}. Emanuel et al. studied the effect of activation functions on CNNs in the context of text classification and found that bounded activation functions, like Sigmoid-based functions, outperform rectifier-based activation functions, like ReLU, in this task ~\cite{EmanuelCNN}.

To date, the effect of activation functions on fNIRS classification tasks remains underexplored and has not been systematically studied in existing literature. However, several studies have explored different deep learning (DL) architectures for classifying fNIRS signals across a range of cognitive and physiological tasks. Table~\ref{tab:summary} provides a summary of some of the recent works applying deep learning models to fNIRS data, detailing the model types and activation functions employed. These models typically rely on standard rectifier-based functions, and relatively few studies have systematically assessed the importance of different activations— partly because ReLU has long been the default in modern CNNs due to its simplicity and effectiveness. For example, Tanveer et al. applied CNN and Multilayer Perceptron (MLP) models for drowsiness detection, employing ReLU and Sigmoid activations \cite{Tanveer}. Similarly, Mirbagheri et al. used a 1D-CNN architecture for stress classification and adopted ELU and Sigmoid \cite{Mirbagheri}. Dargazany et al. developed a dense MLP on raw EEG and fNIRS for emotion recognition, using Leaky ReLU and Softmax for activation \cite{Dargazany}. Saadati et al. trained a CNN to detect emotional states, using ReLU for the hidden layers and Softmax for the classification \cite{Saadati}. Yang et al. utilized a CNN trained on fNIRS-derived features to detect mild cognitive impairment, with ReLU and Softmax activations \cite{Yang}. 
Ghonchi et al. designed a Recurrent-Convolutional Neural Network model using hybrid CNN-Long Short-Term Memory (LSTM) for EEG-fNIRS classification, employing ReLU and Softmax \cite{Ghonchi}.  Ma et al. applied 1D-CNN, which is a CNN with a 1 dimensional kernel, for motor imagery classification, using ReLU and Softmax \cite{Ma}. He et al. proposed MDNN, a multimodal deep neural network designed to integrate EEG and fNIRS signals for classifying cognitive workload. In their fNIRS-only configuration, they utilized ELU for the hidden layers and Softmax for classification head~\cite{HeMDNN}.

\begin{table*}[t]
\centering
\caption{Summary of activation functions used in deep learning models for fNIRS data classification.}
\label{tab:summary}
\begin{tabular}{l l l}
\textbf{Authors (Year)} & \textbf{Model Type} & \textbf{Activation Functions} \\ \hline
\\[-0.8em]
Tanveer et al. (2019) & CNN, MLP & ReLU (hidden), Sigmoid (output) \\ 
Mirbagheri et al. (2019) & 1D-CNN & Likely ReLU (hidden), Sigmoid (output) \\ 
Dargazany et al. (2019) & MLP & Leaky ReLU (hidden), Softmax (output) \\ 
Saadati et al. (2019) & CNN & ReLU (hidden), Softmax (output) \\ 
Ghonchi et al. (2020) & RCNN & ReLU (hidden), Softmax (output) \\
Yang et al. (2020) & CNN & ReLU (hidden), Softmax (output) \\ 
Ma et al. (2021) & 1D-CNN & Likely ReLU (hidden), Softmax (output) \\ 
He et al. (2022) & CNN & ReLU (hidden), Softmax (output) \\ 
Pandey et al. (2024) & CNN & Square, Log (hidden), Softmax (output) \\ 
Adeli et al. (2025) & CNN & Square, Abs, Log (hidden), Softmax (output) \\ \hline
\\[-0.8em]
\end{tabular}
\end{table*}

In the last year however, Pandey et al. introduced fNIRSNet, a dual-branch spatio-temporal CNN designed to classify auditory stimulus-evoked hemodynamic responses, using a mixture of square and Logarithmic activation functions ~\cite{fNIRSNet}. Adeli et al. introduced, AbsoluteNet, a dual brach spatio-temporal CNN with a mixture of Square, Absolute, and Logarithmic activation function ~\cite{absolutenet}. ReLU remains however, the dominant activation function across hidden layers due to its effectiveness in various network architectures applied to fNIRS classification.

In this study, we aim to investigate the effect of different activation functions, namely ReLU, ELU, Swish, Sigmoid, Tanh, Square, and Absolute on the  hemodynamic signal classification using different network architectures. The network architectures, include fNIRSNet ~\cite{fNIRSNet}, AbsoulteNet~\cite{absolutenet}, MDNN~\cite{HeMDNN} as the domain specifics and ShallowConvNet~\cite{Shallow} as the baseline CNN model.  The role of different activation functions in the adopted neural networks was evaluated in the context of auditory processing, using a unified pipeline and consistent hyperparameters across all activation function variants for each model. This comparative study provides critical insights into the role of activation function design in optimizing fNIRS-based DL models for neural decoding tasks. The outcomes can shed light on the importance of customized domain specific activation functions for an enhanced fNIRS classification performance.

\section{ACTIVATION FORMULATION}

\begin{figure*}[t]
  \centering
  \includegraphics[width=\textwidth]{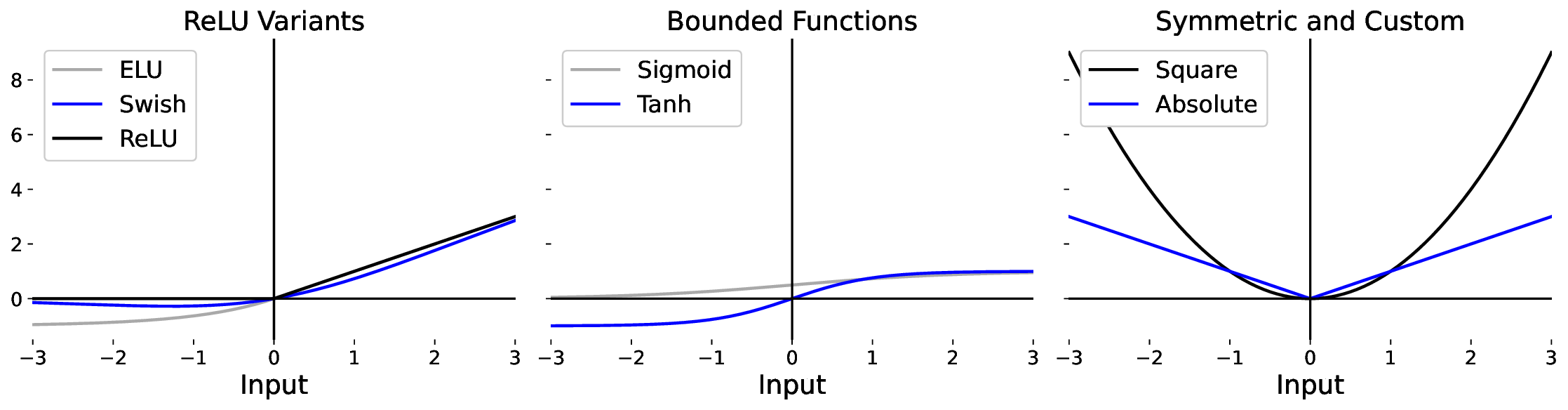}
  \caption{Grouped activation functions used in the study. From left to right: ReLU and its variants (ELU and Swish); bounded activations (Sigmoid, Tanh); and symmetric or custom functions (Square, Absolute).}
  \label{fig:activation_groups}
\end{figure*}

Activation functions introduce non-linearity into neural networks, enabling them to learn complex patterns. Based on the literature review, we chose seven different activation functions, each described mathematically and briefly introduced below. Fig. ~\ref{fig:activation_groups} depicts the activation functions while table ~\ref{tab:actprop} summarizes some of the properties of the functions, which are explained here:

\subsection{Properties}
\begin{itemize}
    \item \textbf{Symmetry}: A function is symmetrical (even) if \( f(x) = f(-x) \), meaning it is mirrored around the \( y \)-axis or (odd) if \( f(x) = -f(-x) \), meaning it is mirrored around the origin. This can be beneficial when modeling data centered around zero.
    
    \item \textbf{Boundedness}: A function is bounded if its output remains within a finite range. Bounded activation functions help prevent exploding activations and can improve stability.

    \item \textbf{Monotonicity}: A monotonic function either never increases or never decreases. Monotonic functions can simplify the optimization landscape during training.

    \item \textbf{Smoothness}: A smooth function is continuously differentiable, which facilitates gradient-based optimization and avoids sharp transitions.
\end{itemize}

\begin{table*}[t]
\centering
\caption{Properties of Activation Functions}
\label{tab:actprop}
\begin{tabular}{ l c c c c c}
\hline
\textbf{Activation Function} & \textbf{Parametric} & \textbf{Monotonic} & \textbf{Smooth} & \textbf{Bounded} & \textbf{Symmetrical} \\
\hline
ReLU        & No  & Yes & No  & No  & No  \\
ELU         & Yes & Yes & Yes & No  & No  \\
Swish       & No  & No  & Yes & No  & No  \\
Simoid      & No  & Yes & Yes & Yes & No  \\
Tanh        & No  & Yes & Yes & Yes & Yes \\
Square      & No  & No  & Yes & No  & Yes \\
Absolute    & No  & Yes & No  & No  & Yes \\

\hline
\end{tabular}
\begin{flushleft}
\centering\textsuperscript{*}\footnotesize{Bounded in implementation due to input clipping, but unbounded in mathematical definition.}
\end{flushleft}
\end{table*}

\subsection{Mathematical Formulation}
\label{subsec:acts}

\begin{itemize}
    \item \textbf{ReLU (Rectified Linear Unit):}  
    ReLU is a widely used activation function that outputs the input if the input is positive; otherwise, it outputs zero. It is computationally efficient and helps mitigate the vanishing gradient problem.
    \begin{equation}
    \text{ReLU}(x) = \max(0, x)
    \end{equation}
    
    \item \textbf{Sigmoid:}  
    The Sigmoid function is a bounded, smooth, and differentiable activation function that maps inputs to the range (0, 1). It is commonly used in binary classification tasks.
    \begin{equation}
    \text{Sigmoid}(x) = \frac{1}{1 + e^{-x}}
    \end{equation}
    
    \item \textbf{Swish:}  
    Swish is a smooth, non-monotonic function proposed by Google researchers ~\cite{swish}. It tends to outperform ReLU in deep models.
    \begin{equation}
    \text{Swish}(x) = x \cdot \sigma(x) = \frac{x}{1 + e^{-x}}
    \end{equation}
    where $\sigma(x)$ is the Sigmoid function.

    \item \textbf{Tanh (Hyperbolic Tangent):}  
    The Tanh function outputs values between -1 and 1. It is zero-centered and useful for models where data is centered around zero.
    \begin{equation}
    \tanh(x) = \frac{e^x - e^{-x}}{e^x + e^{-x}}
    \end{equation}

    \item \textbf{ELU (Exponential Linear Unit):}  
    "ELU improves learning characteristics by allowing negative outputs, which push the mean activations closer to zero and speed up learning.
    \begin{equation}
    \text{ELU}(x) = 
    \begin{cases}
        x & \text{if } x \geq 0 \\
        \alpha(e^x - 1) & \text{if } x < 0
    \end{cases}
    \end{equation}
    where $\alpha$ is a hyperparameter.

    \item \textbf{Absolute Function:}  
    This function outputs the magnitude of the input. While not commonly used as an activation function, it is sometimes employed in feature transformations.
    \begin{equation}
    \text{Abs}(x) = |x|
    \end{equation}
    
    \item \textbf{Square Function:}  
    Squaring the input is a simple transformation used in specific feature engineering contexts.
    \begin{equation}
    \text{Square}(x) = x^2
    \end{equation}

    \item \textbf{Modified Absolute Function:}
   This function resembles the Absolute value function but introduces a scaling factor for negative inputs.
    \begin{equation}
    \text{MAF}(x) = \begin{cases}
        x & \text{if } x \geq 0 \\
        \alpha(x) & \text{if } x < 0
    \end{cases}
    \end{equation}
\noindent
where \(\alpha\) is the scaling factor for the negative side. Here, we chose four different \(\alpha\) values, -2, -1, 0, 2, where \(\alpha = -1\) and \(\alpha = 0\) resembles the Absolute and ReLU functions,  respectively. \(\alpha = -2\) is similar to  \(\alpha = -1\) with a more scaled negative side, and \(\alpha = 2\) is the scaling of the negative side without changing the sign.

\end{itemize}

\section{NETWORK ARCHITECTURES AND DATASET}

\subsection{Networks and Evaluation Process}
The role of different activation functions in several neural networks, including fNIRSNet ~\cite{fNIRSNet}, AbsoluteNet~\cite{absolutenet}, MDNN ~\cite{HeMDNN}, and ShallowConvNet ~\cite{Shallow} was evaluated on an auditory dataset explained in section III.B. fNIRSNet and AbsoluteNet follow a dual-branch spatio-temporal CNN design to extract spatial-temporal and temporal-spatial features, which are then fused using deeper convolutional blocks. In contrast, MDNN adopts a deeper but single-branch spatial-temporal structure, using two convolutional layers for spatial aggregation followed by two additional layers for temporal modeling before reaching the fully connected classification head. Lastly, we included ShallowConvNet, which incorporates a single temporal-spatial convolution block. All these networks apply some form of pruning at the end, using average pooling and/or dropout layers before the final dense layer. For a fair comparison, all output layers employed a Softmax activation function, while hidden layers were configured with the respective activation functions under investigation. A learning rate of \( 9 \times 10^{-4} \) and the Adam optimizer were used consistently across all models, with a batch size of 16. Prior to training, training, validation, and test data were separately standardized over each channel using the MNE Python package. A 5-fold cross-validation (CV) strategy was adopted, in which, for each fold, the data were split into 60\%-20\%-20\% for training, validation, and testing, respectively. Each model was trained for 200 epochs, and the model with the lowest validation loss was selected for an additional 100 epochs using the combined training and validation set (80\%). Results were then reported as the mean and standard deviation across all CV folds.

\subsection{Dataset} 
\label{subsec:dataset}

Functional near-infrared spectroscopy (fNIRS) data were collected during an auditory oddball task using a NIRScout system (NIRx Inc.) with a sampling rate of $f_s = 7.8125$ Hz. 14 channels were recorded from seven sources and eight detectors positioned over the frontal, left auditory, and right auditory cortices. Nine healthy adults (four female, age: $28.33 \pm 9.81$ years) with normal hearing participated in a single session comprising six runs each. The task involved distinguishing deviant tones from standard stimuli. Each run included 20 deviant and 120–140 standard tones, with a 2,000 ms inter-stimulus interval (ISI). This resulted in 1,080 unique deviant trials (9 subjects x 6 runs x 20). Signals were preprocessed by converting light intensity to optical density, applying a 0.005–0.7 Hz bandpass filter, and computing oxygenated (HbO$_2$) and deoxygenated (HbR) hemoglobin concentration via the modified Beer–Lambert Law. Data were upsampled to 10 Hz (for other study purposes), segmented into 15-second epochs post-stimulus (150 time points), and visually inspected for artifacts. A total of 918 clean trials for the deviant class were retained across subjects. To keep the dataset balanced for model training, standard trials were randomly subsampled to match the deviant class. More detail on the dataset is explained in \cite{McLinden}. The total size of the dataset was (1836, 28, 150) (1,836 = 918 x 2 for the two classes, 28 = 14 HbO$_2$ + 14 HbR channels, and 150 = 10 samples/sec x 15 sec).

\section{RESULTS}

The results are presented in tables \ref{tab:fNIRSNet}--\ref{tab:maf}, which report performance metrics (mean ± std) for various activation functions across different neural network architectures. Metrics include training and test accuracy, sensitivity, and specificity. Each table highlights the top-performing activation functions and contextualizes them based on their mathematical properties and their suitability for fNIRS data.

\begin{table*}[t]
\centering
\renewcommand{\arraystretch}{1.2}
\caption{FNIRSNet performance metrics (mean $\pm$ std) for different activation functions. Bold values indicate the highest in each metric.}
\label{tab:fNIRSNet}

\begin{tabular}{lcccc}
\hline
\textbf{AF} & \textbf{Train Accuracy (\%)} & \textbf{Test Accuracy (\%)} & \textbf{Sensitivity (\%)} & \textbf{Specificity (\%)} \\
\hline
ReLU & 99.39 $\pm$ 0.27 & 78.30 $\pm$ 3.95 & 78.75 $\pm$ 5.79 & 77.86 $\pm$ 5.21 \\
ELU & 97.21 $\pm$ 4.58 & 72.49 $\pm$ 4.41 & 71.07 $\pm$ 4.91 & 73.92 $\pm$ 6.04 \\
Swish & 97.88 $\pm$ 2.27 & 72.66 $\pm$ 1.79 & 71.14 $\pm$ 5.45 & 74.18 $\pm$ 4.83 \\
Sigmoid & 98.84 $\pm$ 0.87 & 83.39 $\pm$ 3.04 & 80.94 $\pm$ 5.16 & 85.84 $\pm$ 5.12 \\
Tanh & \textbf{99.22 $\pm$ 0.69} & \textbf{83.66 $\pm$ 5.07} & \textbf{79.73 $\pm$ 5.57} & \textbf{87.59 $\pm$ 4.75} \\
Square & 97.11 $\pm$ 5.94 & 75.93 $\pm$ 7.87 & 77.67 $\pm$ 6.80 & 74.18 $\pm$ 9.63 \\
Absolute & 98.75 $\pm$ 1.08 & 79.00 $\pm$ 2.31 & 79.58 $\pm$ 4.46 & 78.43 $\pm$ 3.58 \\

\hline
\end{tabular}
\end{table*}

\begin{table*}[t]

\centering
\caption{AbsoluteNet performance metrics (mean $\pm$ std) for different activation functions (in percentage). Bold values indicate the highest in each metric.}
\label{tab:Absolutenet}
\renewcommand{\arraystretch}{1.2}
\begin{tabular}{lcccc}
\hline
\textbf{AF} & \textbf{Train Accuracy (\%)} & \textbf{Test Accuracy (\%)} & \textbf{Sensitivity (\%)} & \textbf{Specificity (\%)} \\
\hline
ReLU & 98.28 $\pm$ 1.15 & 75.76 $\pm$ 2.92 & 75.92 $\pm$ 4.73 & 75.60 $\pm$ 2.86 \\
ELU & 96.75 $\pm$ 2.29 & 72.22 $\pm$ 4.83 & 72.55 $\pm$ 2.33 & 71.91 $\pm$ 10.11 \\
Swish & 95.22 $\pm$ 6.70 & 70.69 $\pm$ 8.79 & 69.37 $\pm$ 6.81 & 72.02 $\pm$ 10.89 \\
Sigmoid & 86.75 $\pm$ 6.73 & 66.99 $\pm$ 7.84 & 72.88 $\pm$ 12.48 & 61.12 $\pm$ 17.68 \\
Tanh & \textbf{99.56 $\pm$ 0.23} & \textbf{86.33 $\pm$ 2.10} & \textbf{81.48 $\pm$ 3.45} & \textbf{91.18 $\pm$ 0.96} \\
Square & 93.33 $\pm$ 9.41 & 69.55 $\pm$ 8.26 & 68.96 $\pm$ 9.75 & 70.14 $\pm$ 9.12 \\
Absolute & 98.83 $\pm$ 0.83 & 77.73 $\pm$ 4.73 & 77.13 $\pm$ 5.54 & 78.32 $\pm$ 4.18 \\
\hline
\end{tabular}
\end{table*}

\begin{table*}[t]

\caption{MDNN performance metrics (mean $\pm$ std) for different activation functions. Bold values indicate the highest in each metric.}
\label{tab:MDNN}
\centering
\renewcommand{\arraystretch}{1.2}
\begin{tabular}{lcccc}
\hline
\textbf{AF} & \textbf{Train Accuracy (\%)} & \textbf{Test Accuracy (\%)} & \textbf{Sensitivity (\%)} & \textbf{Specificity (\%)} \\
\hline
ReLU & 98.46 $\pm$ 1.32 & 73.96 $\pm$ 4.13 & 73.10 $\pm$ 7.11 & 74.84 $\pm$ 8.13 \\
ELU & 94.72 $\pm$ 5.22 & 67.38 $\pm$ 6.49 & 68.74 $\pm$ 5.20 & 66.02 $\pm$ 8.45 \\
Swish & 98.01 $\pm$ 1.33 & 72.39 $\pm$ 1.89 & 69.50 $\pm$ 4.38 & 75.26 $\pm$ 4.73 \\
Sigmoid & 98.54 $\pm$ 1.36 & 74.40 $\pm$ 6.82 & 73.43 $\pm$ 12.31 & 75.39 $\pm$ 10.89 \\
Tanh & 97.78 $\pm$ 1.83 & 74.94 $\pm$ 3.27 & 72.87 $\pm$ 2.75 & 77.02 $\pm$ 4.10 \\
Square & 55.51 $\pm$ 3.53 & 50.87 $\pm$ 1.15 & 60.33 $\pm$ 46.94 & 41.31 $\pm$ 47.63 \\
Absolute & \textbf{99.07 $\pm$ 0.75} & \textbf{78.92 $\pm$ 2.56} & \textbf{78.32 $\pm$ 2.67} & \textbf{79.52 $\pm$ 3.80} \\
\hline
\end{tabular}
\end{table*}

\textit{fNIRSNet}: Table~\ref{tab:fNIRSNet} shows that Tanh and Sigmoid achieved the highest test accuracy (83.66\% ± 5.07\%) and (83.39\% ± 3.04\%) respectively and strong scores across all other performance metrics. Both functions are smooth and monotonic, validating their generalizability across architectures. The Absolute function, which is symmetric and monotonic, closely followed (within the standard deviation), underscoring its reliability in deeper networks. ELU and Swish showed the lowest-performing activations with (72.49\% ± 4.41\%) and (72.66\% ± 1.79\%) respectively.

\textit{AbsoluteNet}: Table~\ref{tab:Absolutenet} shows that Tanh achieved the highest performance across all metrics, with a test accuracy of (86.33\%~$\pm$~2.10\%), sensitivity of (81.48\%~$\pm$~3.45\%), and specificity of (91.18\%~$\pm$~0.96\%). This highlights its suitability for deeper architectures like AbsoluteNet, where gradient flow and saturation behavior play a larger role. The Absolute function also showed strong and stable results, with high training accuracy and competitive test performance (77.73\%~$\pm$~4.73\%), confirming its robustness across architectures. In contrast, Sigmoid, which had excelled in fNIRSNet, underperformed in AbsoluteNet, showing the lowest training accuracy (86.75\%~$\pm$~6.73\%). ReLU, offered balanced results and outperformed ELU, Swish, Square and Sigmoid.

\textit{MDNN}: In table~\ref{tab:MDNN}, the Absolute function delivered the highest test accuracy (79.92\% ± 2.56\%) and outperformed all other functions in sensitivity and specificity as well. Its symmetry and simplicity likely help it generalize well without overfitting. Tanh and Sigmoid achieved the second highest training accuracy and the second highest test accuracy of (74.94\% ± 3.27\%) and (74.40\% ± 6.82\%), respectively. Other typical functions including ReLU, Swish performed moderately, while ELU and Square underperformed.

\begin{table*}[t]

\caption{ShallowConvNet Network performance metrics (mean $\pm$ std) for different activation functions. Bold values indicate the highest in each metric.}
\label{tab:shallow}
\centering
\renewcommand{\arraystretch}{1.2}
\begin{tabular}{lcccc}
\hline
\textbf{AF} & \textbf{Train Accuracy (\%)} & \textbf{Test Accuracy (\%)} & \textbf{Sensitivity (\%)} & \textbf{Specificity (\%)} \\
\hline
ReLU & 89.17 $\pm$ 2.86 & 66.56 $\pm$ 2.74 & 67.32 $\pm$ 5.11 & 65.80 $\pm$ 3.57 \\
ELU & 78.72 $\pm$ 8.40 & 64.38 $\pm$ 3.18 & 64.16 $\pm$ 5.59 & 64.60 $\pm$ 1.36 \\
Swish & 83.35 $\pm$ 7.01 & 64.05 $\pm$ 3.13 & 62.31 $\pm$ 6.24 & 65.79 $\pm$ 6.47 \\
Sigmoid & 68.29 $\pm$ 1.11 & 60.08 $\pm$ 3.56 & 63.95 $\pm$ 6.37 & 56.23 $\pm$ 10.06 \\
Tanh & \textbf{91.86 $\pm$ 2.55} & \textbf{68.63 $\pm$ 2.81} & \textbf{67.31 $\pm$ 4.45} & \textbf{69.94 $\pm$ 3.07} \\
Square & 84.27 $\pm$ 9.54 & 57.79 $\pm$ 1.64 & 58.82 $\pm$ 8.74 & 56.74 $\pm$ 9.11 \\
Absolute & 91.94 $\pm$ 2.76 & 64.98 $\pm$ 4.25 & 64.38 $\pm$ 8.00 & 65.57 $\pm$ 10.02 \\
\hline
\end{tabular}
\end{table*}

\begin{table*}[t]
\label{tab:maf}
\centering
\caption{Performance metrics (mean $\pm$ std) for different alpha values (all values in \%)}
\begin{tabular}{c c c c c}

\textbf{Alpha} & \textbf{Train Accuracy (\%)} & \textbf{Test Accuracy (\%)} & \textbf{Sensitivity (\%)} & \textbf{Specificity (\%)} \\
\hline
\\[-0.8em]
\multicolumn{5}{c}{\textbf{MDNN}} \\
\hline
\\[-0.8em]
-2    & 96.80 $\pm$ 7.48 & 76.47 $\pm$ 2.84 & 75.50 $\pm$ 10.30 & 77.45 $\pm$ 6.72 \\
-1    & \textbf{98.03 $\pm$ 1.47} & \textbf{76.53 $\pm$ 2.84} & \textbf{76.58 $\pm$ 6.26} & \textbf{76.47 $\pm$ 8.36} \\
0     & 97.33 $\pm$ 2.92 & 72.17 $\pm$ 5.57 & 70.87 $\pm$ 3.39 & 73.49 $\pm$ 9.07 \\
2     & 89.85 $\pm$ 3.12 & 64.95 $\pm$ 64.95 & 66.07 $\pm$ 3.80 & 63.84 $\pm$ 6.69 \\
\\[-0.8em]
\multicolumn{5}{c}{\textbf{AbsoluteNet}} \\
\hline
\\[-0.8em]
-2 & 98.78 $\pm$ 1.15 & 83.82 $\pm$ 3.18 & 82.19 $\pm$ 4.09 & 85.45 $\pm$ 4.22 \\
-1 & \textbf{99.60 $\pm$ 0.50} & \textbf{85.70 $\pm$ 2.75} & \textbf{83.40 $\pm$ 2.11} & \textbf{88.20 $\pm$ 3.72} \\
0  & 68.76 $\pm$ 14.68 & 55.94 $\pm$ 8.44 & 56.97 $\pm$ 9.71 & 54.90 $\pm$ 13.36 \\
2  & 62.19 $\pm$ 7.10 & 53.87 $\pm$ 3.17 & 51.65 $\pm$ 8.61 & 56.10 $\pm$ 10.29 \\
\\[-0.8em]
\hline
\end{tabular}
\end{table*}

\textit{ShallowConvNet}: In table~\ref{tab:shallow}, Tanh again performed best in terms of test accuracy (68.63\% ± 2.81\%) as well as in other metrics. Absolute function (64.98\% ± 4.25\%) along with the ReLU (66.56\% ± 2.74\%), ELU (64.38\% ± 3.18\%), and Swish (64.05\% ± 3.13\%) all showed stable, balanced performance across all metrics. However, the Absolute function showed much higher training accuracy (91.94\% ± 2.76\%) in comparison to the ELU (78.72\% ± 8.40\%) and Swish (83.35\% ± 7.10\%) functions. This might be a sign of the Absolute function having the ability to capture hemodynamic response features from the fNIRS signals. The low test accuracy can be attributed to the shallow and compact design of the network, hence the name, and the network not being fNIRS-specific. As with other models, Square ranked lowest, confirming its unsuitability, which may require more tuning before widespread implementation.

\textit{}{\textit{Modified Absolute Function (MAF)}}: Table~\ref{tab:maf} presents a controlled evaluation of the Modified Absolute Function (MAF) for different values of the parameter \(\alpha\) on two of the networks (AbsoluteNet and MDNN), allowing us to isolate the impact of symmetry and non-linearity within a single function family. As observed, \(\alpha = -1\) outperformed other \(\alpha\) values and showed stable performance across both networks. These results suggest that symmetry in the activation function—treating positive and negative inputs equally—might be beneficial in modeling fNIRS signals, which potentially contain meaningful information in both polarities. \(\alpha = -2\) also showed  promising results, comparable to \(\alpha = -1\) in MDNN and closely following \(\alpha = -1\) in AbsoluteNet. Conversely, when \(\alpha = 0\), ReLU, the performance was consistently lower, indicating that ignoring the negative component of the weighted sum of each neuron may limit the model’s representational power for fNIRS data. Furthermore, larger positive values of \(\alpha\), such as \(\alpha = 2\), degrades the performance and in the case of AbsoluteNet shows chance-level results across test accuracy, sensitivity, and specificity proving the importance symmetry in the shape of the activation function.
Overall, the results suggest that activation functions with a positive output to negative input— like Absolute—can outperform widely used rectified-based activations such as ReLU. This reinforces the importance of activation symmetry in fNIRS classification applications, where both positive and negative neuron output can potentially help the network extract features related to hemodynamic responses.

\section{CONCLUSIONS}

Overall, our results demonstrate the critical role of activation functions in improving classification performance for deep neural networks in fNIRS-based classification tasks. Across all tested activation functions, Tanh and Absolute consistently ranked among the top-performing activation functions in all of the tested neural networks. These functions share key properties such as symmetry and, in the case of Tanh, smoothness and boundedness. These attributes might especially be well-suited for fNIRS data, which contain meaningful bidirectional signal variations. By contrast, the commonly used ReLU was outperformed in most cases, and the Square function, when used alone, consistently yielded the weakest results. It is worth noting that Absolute function showed its strongest performance in deeper architectures such as MDNN and AbsoluteNet, but was less effective in shallower networks (ShallowConvNet). We may attribute the superior results of the Absolute activation function to its, relative simple design, with non-zero output for the negative inputs and its symmetrical shape. Symmetry may help preserve meaningful polarity information in fNIRS signals, which exhibit bidirectional variations around a mean.

The targeted MAF analysis further confirmed that symmetrical activations like Absolute outperform ReLU when tested under otherwise identical conditions. This emphasizes the importance of aligning activation function properties with domain-specific signal characteristics. These results demonstrate that carefully selecting activation functions based on data properties—rather than defaulting to typical rectified-based choices—may enhance the performance of deep learning models in fNIRS classification tasks.

This study faces two limitations. First, the current analysis investigates the impact of activation functions by uniformly replacing all activation layers with a single type, thereby limiting the exploration to homogenous activation strategies. This approach, while informative, overlooks the potential advantages of heterogeneous activation function architectures—where different layers or subcomponents of the network employ distinct activation functions tailored to their functional roles. Future research should systematically explore these hybrid activation designs, investigating how specific combinations—conditioned on layer depth, network topology, or task-specific characteristics—affect model performance, convergence behavior, and robustness. Second, testing on only one dataset pertaining to auditory processing with a rather limited number of samples may constrain the generalizability of our findings to other tasks and datasets. Future work will address this by evaluating the network performance on multiple datasets. 

\addtolength{\textheight}{-12cm}   




\section*{ACKNOWLEDGMENT}

This work was supported by the National Science Foundation [NSF-2024418, NSF-2417447].

\IEEEtriggeratref{21}

\end{document}